%% file: main.tex
\documentclass[runningheads]{llncs}
\usepackage[T1]{fontenc}
\usepackage{booktabs}
\usepackage{graphicx}
\usepackage{wrapfig}
\usepackage{lscape}
\usepackage{rotating}
\usepackage{todonotes}
\usepackage[dvipsnames]{xcolor}
\usepackage[normalem]{ulem}
\usepackage{threeparttable}

\usepackage{listings}
\usepackage[table]{xcolor}
\usepackage{enumitem}
\usepackage{amssymb}
\usepackage{tabularx}
\lstdefinelanguage{SPARQL}{
  morekeywords={SELECT,WHERE,PREFIX,FILTER,OPTIONAL,a},
  sensitive=true,
  morecomment=[l]{\#},
  morestring=[b]"
}

\begin{document}
\title{The AnIML Ontology: Enabling Semantic Interoperability for Large-Scale Experimental Data in Interconnected Scientific Labs}
% \title{IDEA2: A Collaborative Expert-in-the-Loop Workflow for Competency Question Engineering using Large Language Models}
%
\titlerunning{The AnIML Ontology}
% If the paper title is too long for the running head, you can set an abbreviated paper title here
%
% [AUTHOR LIST: hidden for double-blind]
\author{Wilf Morlidge\inst{1} \and
Elliott Watkiss-Leek\inst{1}\orcidID{0009-0002-9284-6590} \and
George Hannah\inst{1}\orcidID{0000-0002-3218-4559} \and
Harry Rostron\inst{2} \and
Andrew Ng\inst{2} \and
Ewan Johnson\inst{2} \and
Andrew Mitchell\inst{2} \and
Terry R. Payne\inst{1}\orcidID{0000-0002-0106-8731} \and
Valentina Tamma\inst{1}\orcidID{0000-0002-1320-610X} \and
Jacopo de Berardinis\inst{1}\orcidID{0000-0001-6770-1969}}

\authorrunning{W. Morlidge et al.}
% \authorrunning{Anonymous et al.}
% [BELOW: hidden for double-blind]
\institute{School of Computer Science \& Informatics, University of Liverpool, UK \\
Unilever Plc. Materials Innovation Factory, University of Liverpool, UK \\
\email{\{jacodb,v.tamma\}@liverpool.ac.uk}}
\maketitle              % typeset the header of the contribution
\begin{abstract}
Achieving semantic interoperability across heterogeneous experimental data systems remains a major barrier to data-driven scientific discovery.
The Analytical Information Markup Language (AnIML), a flexible XML-based standard for analytical chemistry and biology, is increasingly used in industrial R\&D labs for managing and exchanging experimental data.
However, the expressivity of the XML schema permits divergent interpretations across stakeholders, introducing inconsistencies that undermine the interoperability the AnIML schema was designed to support.
In this paper, we present the \textbf{AnIML Ontology}, an OWL\,2 ontology that formalises the semantics of AnIML and aligns it with the Allotrope Data Format to support future cross-system and cross-lab interoperability.
The ontology was developed using an expert-in-the-loop approach combining LLM-assisted requirement elicitation with collaborative ontology engineering.
We validate the ontology through a multi-layered approach: data-driven transformation of real-world AnIML files into knowledge graphs, competency question verification via SPARQL, and a novel validation protocol based on adversarial negative competency questions mapped to established ontological anti-patterns and enforced via SHACL constraints.
%We also release competency questions, examples, documentation, alignments, and supporting material to facilitate the reuse and the extension of the ontology.
\end{abstract}

\input{sections/introduction}

\input{sections/related}
\input{sections/ontology_development}
\input{sections/ontology}
\input{sections/discussion_conclusions}

\section*{Acknowledgements}
This work was supported by the Royal Academy of Engineering under the Google DeepMind Research Ready Scheme, and by an EPSRC ICASE studentship (201146) with Unilever PLC.
We would like to thank Diego Conde Herreros for his valuable assistance with the SSSOM mapping definitions.
% \newpage
\bibliographystyle{plain}
\bibliography{references}

\end{document}

%% file: sections/introduction.tex
\section{Introduction}

Advances in scientific research has resulted in increasingly large volumes of distributed experimental data.  Although interoperability (i.e. the ability of systems to exchange and consistently interpret data) is essential for the aggregation of such data and its subsequent dissemination to geographically dispersed collaborators~\cite{anderson2025looking}, the diversity of analytical platforms, software systems, and data recording conventions produces significant variability in how experimental information is structured and interpreted~\cite{bai2024dynamic}.
%, this undermining data reuse, and automation.
%Scientific research increasingly generates vast experimental data across distributed laboratories, instruments, and teams~\cite{anderson2025looking}. The diversity of analytical platforms, software systems, and data recording conventions produces significant variability in how experimental information is structured and interpreted~\cite{bai2024dynamic}, making semantic interoperability, i.e. the ability of systems to exchange and consistently interpret data, a persistent barrier to data reuse, integration, and automation.

Syntactic standards have traditionally been used to address this interoperability challenge.
The \textit{Analytical Information Markup Language
(AnIML)}~\cite{schafer2004documenting} is one such XML-based format for encoding experimental data in analytical chemistry and biology, consisting of a core schema and technique-specific definition documents
(ATDDs)~\cite{schaefer20XXschema}.
However, the flexibility afforded by the XML schema permits divergent interpretations across stakeholders: within the same organisation, laboratories may adopt differing practices when populating AnIML files,
producing subtle but critical mismatches in terminology, structure, and metadata~\cite{doan2001reconciling}.
These inconsistencies undermine efforts to federate data across systems and to align AnIML with complementary frameworks such as the \textit{Allotrope Data
Format}~\cite{gardiner2024rise}.
%The problem intensifies as
The need for solutions that facilitate interoperability have become increasingly urgent with the growth in numbers of 
autonomous laboratories, where robotic systems generate data with minimal human intervention, place new demands on data harmonisation and automated reasoning~\cite{bai2024dynamic,stach2021autonomous}.

% Prior work has investigated deriving ontologies directly from XML schemas
% through automated transformation rules~\cite{Bedini2010,bohring2005mapping}.
% However, these approaches assume a relatively straightforward mapping between
% schema elements and ontological constructs. AnIML resists such treatment:
% its core entities, i.e. \texttt{Category}, \texttt{Parameter}, and
% \texttt{Reference}, are reused across different contexts (samples,
% methods, results, infrastructure) with distinct intended semantics in each,
% making rule-based XSLT transformations unreliable. The underlying difficulty
% is that AnIML was designed not merely as a data exchange format but as an
% implicit conceptual model of analytical experimentation that encodes
% domain requirements in its structure without formalising them explicitly. Our
% approach starts from this observation: if the schema implicitly encodes an
% ontology, its requirements can be made explicit through systematic
% elicitation, and the resulting formal model can resolve the interpretive
% ambiguities that the XML encoding leaves open.

Prior work on deriving ontologies automatically from XML schemas~\cite{Bedini2010,bohring2005mapping}.
has proven unreliable for AnIML: its core entities are reused across
different contexts with distinct intended semantics, resisting rule-based
transformation. The underlying challenge is that AnIML was designed not
just as a data exchange format but as an implicit conceptual model of
analytical experimentation that encodes domain requirements without
formalising them explicitly. 
%Our approach starts 
We from this observation:
if the schema implicitly encodes an ontology, its requirements can be
made explicit through systematic elicitation, 
with a resulting formal model that resolves
%and the resulting formal model can resolve 
the interpretive ambiguities 
inherent in the XML encoding.
%that the XML encoding leaves open.

%Ontologies offer a principled solution. 
As formal, explicit specifications of shared conceptualisations~\cite{studer1998knowledge}, ontologies go beyond syntactic schemas: they can be specialised and extended while maintaining
logical consistency, and alignments between ontologies can bridge independent standards to mitigate semantic heterogeneity.
This approach has proven effective in the life
sciences~\cite{ashburner2000gene,chen_et_al:TGDK.1.1.5,donnelly2006snomed}, and recent work demonstrates that formal semantic layers over experimental data can reduce redundant workflows and data inconsistencies~\cite{coley2020autonomous,gadiya2023fair,wilkinson2016fair}.
% LARASuite~\cite{Doerr_Maak_Menke_Bornscheuer_2023} uses semantic
% technologies for FAIR laboratory automation, while Bai et
% al.~\cite{bai2024dynamic} employ ontologies and a dynamic knowledge graph
% to link distributed self-driving laboratories across institutions.

In this paper, we present the \textbf{AnIML Ontology},\footnote{All resources are available at {\url{https://github.com/KE-UniLiv/animl-ontology}}, under the Creative Commons Attribution 4.0 International (CC-BY 4.0) license.}
an OWL\,2 ontology
that formalises the semantics of AnIML, resolving the ambiguities inherent
in its XML schema by providing precise intended interpretations for its
core concepts and relationships. The ontology also aligns with the
Allotrope Data Format to facilitate cross-standard interoperability. The
contributions of this work are:
\begin{enumerate}[leftmargin=*,nosep]
    \item The AnIML Ontology: a modular OWL\,2 ontology grounded in
established ODPs, based on 102 expert-validated competency questions,
including a new Content ODP (the \emph{AnIML Reference Pattern}) that
formalises implicit data pointers opaque to semantic reasoners.
    \item Scoped ontology alignments with the Allotrope Foundation
    Ontologies, released in SSSOM format, demonstrating how purposive
    (rather than exhaustive) alignment can support cross-standard
    interoperability when source and target differ in modelling
    granularity.
    \item A validation protocol based on adversarial negative competency
    questions, mapped to established ontological
    anti-patterns~\cite{sales2015antipatterns} and enforced via SHACL,
    that complements standard CQ verification by systematically checking for unintended model instances.
\end{enumerate}
%
%All resources are available at {\url{https://github.com/KE-UniLiv/animl-ontology}}, under the Creative Commons Attribution 4.0 International (CC-BY 4.0) license.

%% file: sections/related.tex
\section{Related work}\label{sec:related-work}

Ontologies have experienced significant adoption in the biomedical and life
sciences domains~\cite{rubin2008bioportal}, ranging from broad resources such
as the Systematized Nomenclature of Medicine~\cite{donnelly2006snomed} and the
Gene Ontology~\cite{ashburner2000gene} to targeted models addressing specific
diseases~\cite{he2020cido}. In the chemical domain, the Chemical Information
Ontology~\cite{Hastings2011} and OntoSpecies~\cite{Akroyd2023} provide
frameworks for classifying chemical properties, species, and algorithms, while
large-scale integration projects like Open PHACTS~\cite{williams2012facts}
demonstrate how ontologies can bridge isolated vocabularies into comprehensive
FAIR data ecosystems. A recent survey by Strömert et
al.~\cite{stromert2022ontologies4chem} maps the landscape of chemistry
ontologies (including ChEBI~\cite{dematos2010chebi}, CHMO, and the Allotrope
Foundation Ontologies) relative to BFO and the OBO Foundry, highlighting both
the breadth of the ecosystem and the gaps that remain in formalising
experimental workflows and instrument data.

\noindent\textbf{Knowledge representation for autonomous laboratories.}
A primary motivation for our work is its potential application in autonomous laboratories, where robotic systems execute experiments with minimal human intervention.
This vision requires a rich knowledge representation layer to support automated reasoning, planning, and execution~\cite{aitken2017ras}, a necessity long recognised in autonomous robotics~\cite{Paull2012}.
Recent work demonstrates the feasibility of ontology-driven laboratory automation.
%Bai et al.~\cite{bai2024dynamic} develop a dynamic knowledge graph architecture within The World Avatar project, employing a suite of ontologies --including OntoSpecies for chemical entities and OntoKin for reaction kinetics -- to link distributed self-driving laboratories across institutions, with robots in Cambridge and Singapore conducting collaborative closed-loop optimisation.
%LARASuite~\cite{Doerr_Maak_Menke_Bornscheuer_2023} takes a complementary approach, using semantic technologies to automate data capture and standardise metadata annotation within FAIR experimental workflows.
%More recently, the Chemotion Knowledge Graph~\cite{norouzi2025chemotion} adopts Ontology Design Patterns to ground electronic lab notebook data in BFO, while the Process Chemistry Ontology (PROCO), jointly developed by academia and the Allotrope Foundation, bridges the OBO Foundry and Allotrope ecosystems for process chemistry~\cite{stromert2022ontologies4chem}.
At a higher level of abstraction, the Autonomica project~\cite{Candela2023} formalises the architecture of model-based autonomous systems with a focus on goals, states, planning, and scheduling.
While such frameworks provide a reasoning foundation for autonomous control, they require a rich, domain-specific knowledge base to operate effectively.
% %%%%%%%%%%%%%%%%%%%%%%%%%%%%%%%%%%%%
% %% The following text would work better in the discussion or conclusions
%%%%%%%%%%%%%%%%%%%%%%%%%%%%%%%%%%%%%%
%The AnIML Ontology is designed to provide this knowledge for scientific experimentation, bridging high-level control ontologies and the low-level, domain-specific data they must interpret. Unlike the domain ontologies above, which target specific chemical entities, reactions, or processes, the AnIML Ontology formalises the \emph{structure of experimental records} as defined by an established industrial standard, thus addressing the layer of data organisation that connects instrument output to downstream reasoning.
%%%%%%%%%%%%%%%%%%%%%%%%%%%%%%%%%%%%%%

\noindent\textbf{Ontology engineering from XML schemas.}
XML schemas like AnIML are widely used to define the syntactic structure of data for exchange.
While effective for providing hierarchical structure, the flexibility afforded by XML schemas means that the specification lacks the formal semantics required for robust data integration and machine reasoning~\cite{doan2001reconciling}.
A significant body of research has investigated deriving ontologies from XML data and schemas~\cite{Duan2023_XSD2SHACL}.
Bedini et al.~\cite{Bedini2010} develop a method to derive a ``skeleton ontology'' from XML schema files, and Hannah et al.~\cite{hannah2023towards} generate a skeleton from an XML schema using a rule-based approach subsequently enriched via LLMs; both use the rule set by Bohring and Auer~\cite{bohring2005mapping}.
However, these outputs still require substantial refinement by domain experts~\cite{Bedini2010}, as automated tools can map syntactic structures but cannot reliably capture the implicit domain knowledge encapsulated by a schema.
Recent work shows promise in using LLMs as domain knowledge elicitors to refine skeleton ontologies~\cite{hannah2025large}, potentially reducing expert workload~\cite{hannah2025relrae}, though these contributions remain preliminary and require established ground truth to validate outputs and mitigate hallucinations.
% %%%%%%%%%%%%%%%%%%%%%%%%%%%%%%%%%%%%%%
% %% The following text would work better in the discussion or conclusions
%%%%%%%%%%%%%%%%%%%%%%%%%%%%%%%%%%%%%%
% Our work differs from these semi-automatic approaches: rather than extracting an ontology from the XML schema, we use the schema as a starting point for expert-driven ontology engineering, with the resulting model grounded in Ontology Design Patterns and validated against both real-world data and adversarial competency questions.

%% file: sections/ontology_development.tex
\section{Ontology development process}\label{sec:development}

The development of the AnIML Ontology follows our existing approach
%that our group has been progressively refining 
for building ontologies from existing data
sources in collaboration with industry
stakeholders~\cite{mansfield2021capturing}. The central design principle is
to maximise the accuracy of domain expert contributions while minimising
the time they spend on tasks that do not require their specific expertise.
In prior work~\cite{mansfield2021capturing}, we achieved this by using UML
diagrams and controlled-language glossaries as a \textit{lingua franca}
between ontology engineers and domain experts, enabling stakeholders to
reach consensus during the knowledge capture phase with minimal exposure
to ontology formalisms. In this paper, we extend this principle by
introducing LLMs as a drafting tool for requirement elicitation,
further reducing the expert workload by automating the initial extraction
of candidate requirements from large technical documents.

The process does not follow any single ontology engineering methodology
verbatim, but is grounded in the phase structure common to the principal
approaches in the literature: NeOn~\cite{suarez2011neon}, Ontology
101~\cite{noy2001ontology}, eXtreme Design~\cite{presutti2009extreme}, Methontology~\cite{fernandez1997methontology},
and Pay-As-You-Go~\cite{sequeda2017pay}. These share a common core:
requirements elicitation and competency question formulation,
collaborative modelling and iterative refinement, validation against
structural and logical criteria, and verification against domain
requirements. Our process maintains all of these phases while making
pragmatic tooling choices motivated by the specific conditions of this
project, detailed in the following subsections.
%\footnote{We release all code and resources at \url{https://github.com/KE-UniLiv/animl-ontology}.}.

\subsection{Expert-in-the-loop requirement engineering}
\label{ssec:requirement-engineering}

Standard ontology requirement engineering elicits requirements through interviews or workshops, reorganises them into user stories, and derives competency questions (CQs)~\cite{malone2014software,presutti2009extreme}. 
For AnIML, explicit requirements are unavailable, so we elicit CQs directly from the core and technique definition documents, treating the schema's modelling choices as implicit expressions of its requirements. This creates a knowledge gap: ontology engineers can formalise atomic, verifiable CQs but lack the domain expertise to interpret the schema; domain experts possess deep scientific knowledge but often struggle to formulate formal CQs~\cite{de2023polifonia}. The core schema alone spans approximately 2,500 lines, making manual extraction both slow and error-prone.
We address this by using LLMs as a drafting tool for CQ elicitation — not as an autonomous requirements engineer, but to accelerate extraction of candidate requirements from large technical documents, with all outputs subject to expert validation. This extends our earlier work~\cite{mansfield2021capturing}, where UML served as the intermediary representation between engineers and experts; here the LLM produces an initial draft that experts accept, reject, or refine, preserving the same separation of concerns.

\noindent\textbf{Expert feedback loop.}
Annotators review each CQ, assigning tags and comments, and accept or reject it. Rejection criteria include: failure to address an AnIML requirement, ambiguity, excessive complexity, or direct references to XML schema elements rather than conceptual entities. A majority vote mechanism determines fate: CQs with a net score $\le0$ are flagged for LLM reformulation in the next iteration, using a prompt constructed from the full message history, schemas, and anonymised expert feedback. This decouples elicitation (LLM) from validation (experts) and prevents the propagation of hallucinations.

%During each iteration, annotators review the CQs, assigning thematic tags or priorities, adding comments to express concerns, and suggesting rephrasing. Annotators are instructed to \textit{accept} or \textit{reject} each CQ. Rejection criteria include: failure to address a requirement supported by AnIML, ambiguity, excessive complexity (e.g., entailing multiple requirements), or direct references to specific XML schema elements (e.g., \texttt{ExperimentStep}) rather than conceptual entities. We employ a majority vote mechanism for validation. If a CQ receives a negative net score ($\leq 0$), it is flagged for reformulation in the subsequent iteration. This cycle offers two benefits: (1)~it allows domain experts to filter LLM hallucinations or irrelevant CQs before they propagate into the design; and (2)~it steers the LLM to reformulate CQs based on specific expert feedback, creating a refined prompt loop.

\noindent\textbf{Implementation.}
The generation of CQs is based on a \emph{formulation prompt} following the few-shot approach proposed in~\cite{zhang2024ontochat}, optimised for CQ extraction.
When CQs are flagged for revision, a \emph{reformulation prompt} is constructed from the full message history up to that iteration, comprising the AnIML schemas, the original formulation prompt, and all CQs with their scores, votes, and anonymised expert feedback.
The workflow uses \texttt{Gemini 2.5 Pro} via the Google Cloud API, selected for its large context window, which is necessary to process the verbose AnIML \texttt{core} and \texttt{technique} schemas. Evaluators access CQs through a web-based dashboard that serves as the
crowdsourcing platform.
% populated and controlled via API.

\noindent\textbf{CQ validation.}
The evaluation was performed by four experienced domain experts from
Unilever.
%our industry partner, with 2--10 years of experience using AnIML.
Such expert feedback was critical for refining the model's output; e.g., the initial CQ ``\textit{What
result blueprints are defined for a technique?}'' was rejected by two
annotators for ambiguity. Following the feedback loop, this was
automatically reformulated as ``\textit{What are the names of the result
definitions specified by a technique?}'', which was subsequently accepted
with a score of~$3$.
In total, 102 CQs were validated across 3 iterations.
The low rate of manual intervention (one CQ) reflects the cumulative effect of the iterative feedback loop: CQs that did not meet expert approval were automatically reformulated using the annotators' feedback, with most issues resolved within one or two reformulation cycles. The validation  (Section~\ref{sec:validation_adversarial}) provides a
complementary check on whether these CQs are sufficient to exclude unintended models.

\begin{sidewaysfigure}[p]
    \includegraphics[width=\textwidth]{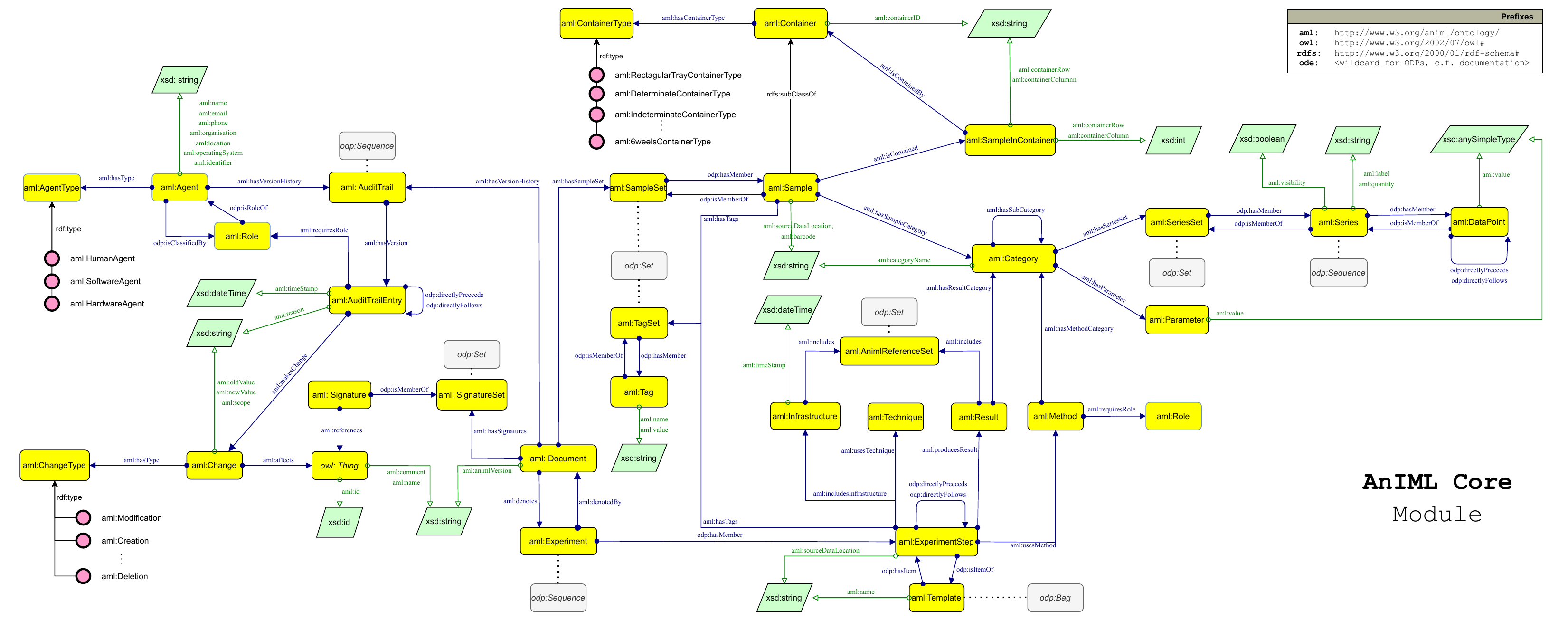}
    \caption{Overview of the \textbf{AnIML Core} module using the Graffoo notation \cite{falco2014modelling}: yellow boxes denote classes, blue/green arrows are object/datatype properties, purple circles are individuals, green polygons are datatypes. We also denote ODP reuse using grey boxes.}
    \label{fig:animl-core}
\end{sidewaysfigure}
\subsection{From CQs to collaborative ontology design}
\label{ssec:ontology-design}

Validated CQs are clustered into thematic groups, which facilitates the identification of common requirements and their mapping to the modular structure of AnIML.
This stage follows the XD approach~\cite{presutti2009extreme}, utilising
Graffoo~\cite{falco2014modelling} to collaboratively construct graphical representations of the ontology in an interdisciplinary setting.
% To promote semantic clarity and extensibility, we prioritise the identification and reuse of Ontology Design Patterns (ODPs) -- modelling solutions to recurrent design problems~\cite{gangemi2009ontology}.
% Their reuse enhances semantic interoperability by grounding the model in established, community-validated
% structures\footnote{\url{ontologydesignpatterns.org}}, ensuring the resulting ontology is semantically robust and easier to maintain.
To promote semantic clarity and extensibility, we prioritise the
identification and reuse of Ontology Design Patterns
(ODPs)~\cite{gangemi2009ontology}. We reuse Content ODPs, such as the \texttt{Set}, \texttt{Sequence}, and \texttt{Bag} patterns to describe collections, and the \texttt{Situation} pattern~\cite{borgo2022dolce} to model measurements and experimental contexts as first-class entities.
We selected these Content ODPs due to their prevalence in life science ontologies~\cite{mortensen2012applications} and their conceptual consistency with alignment targets grounded in the Basic Formal Ontology (BFO) \cite{otte2022bfo}.

The development team comprised two ontology engineers, with two additional engineers focused on validation. Iterative whiteboarding sessions using Graffoo allowed the team to refine module boundaries and enhance the
clarity of the ontology.
This phase consisted of 3 iterations where ontology prototypes were progressively refined to maximise ODP reuse while minimising model complexity (i.e., reducing unnecessary classes and properties).
This also led to the definition of a new ODP (\textit{AnIML
Reference}, c.f.\ Section~\ref{ssec:animl-reference-odp}). Ontology versions were validated against a set of anonymised AnIML files provided by the stakeholder to ensure that the model could accurately cover real-world
data instances.

The resulting ontology is encoded in OWL~2. We selected this representation language because it is the W3C standard for formal knowledge representation~\cite{horrocks2003shiq}, and the target ontologies we need to align with (Allotrope Foundation Ontologies, ChEBI \cite{dematos2010chebi}, and the broader OBO Foundry ecosystem \cite{smith2007obo} are all published in OWL.
Alternative representation languages were considered: conceptual modelling frameworks such as OntoUML \cite{guerson2015ontouml} offer richer ontological distinctions but require a lossy translation step to OWL for deployment in the RDF/SPARQL ecosystem needed for KG population and CQ verification; purely constraint-based approaches such as SHACL-only models can enforce structural validity but cannot express the class-level axioms and alignments needed for cross-standard interoperability; and lightweight vocabularies such as SKOS lack the expressivity to formalise the domain constraints captured by our technique specifications and adversarial validation.
Choosing a different formalism would require a translation layer that introduces exactly the kind of semantic ambiguity our work aims to eliminate.

Additionally, the OWL/RDF stack provides an integrated toolchain (Protégé for editing, SPARQL for CQ verification, SHACL for constraint validation, and reasoners for consistency checking) that supports the full development and evaluation lifecycle.
Finally, annotation properties and domain and range constraints were added to enrich the model's semantics and metadata.
OWL axioms were defined to facilitate the verification of the ontology, while SHACL shapes~\cite{Knublauch2017SHACL} were implemented to enforce structural constraints.
% (e.g., for the AnIML reference pattern).

%% file: sections/ontology.tex
\section{The AnIML ontology}\label{sec:animl-ontology}

The ontology is implemented in the Web Ontology Language (OWL) \cite{horrocks2003shiq} and uses the namespace \texttt{\url{http://www.w3id.org/animl/ontology/}} (prefixed as \texttt{aml}).
We follow the Semantic Versioning format \cite{garijo2020best} (X.Y.Z), where X denotes major versions (significant model changes or new requirements).
% Y denotes minor versions (backward-compatible additions), and Z denotes patches (bug fixes or metadata updates).
\texttt{v1.1.0} comprises $49$ classes, $54$ object properties, $34$ data properties, and $39$ individuals.
The resource is distributed as an open-source project under the Creative Commons Attribution 4.0 International (CC-BY 4.0) license.

The ontology is organised into two modules -- \texttt{core} and \texttt{technique} -- mirroring their separation in AnIML.
We also introduce a new ODP, the \textit{AnIML Reference Pattern}, to formalise the implicit data pointers used within the schema.

\subsection{The core module}\label{ssec:core}

The \texttt{core} module, illustrated in Figure~\ref{fig:animl-core}, formalises the AnIML Core Schema.
It represents the primary structure for experimental data, metadata, and provenance.
In the ontology, this is rooted in the \texttt{aml:Document} class, which serves as the central container aggregating four main components: the experimental data (\texttt{aml:Experiment}), the samples involved (\texttt{aml:SampleSet}), provenance and versioning (\texttt{aml:AuditTrail}), and digital signatures (\texttt{aml:SignatureSet}).

\vspace{-1em} % FIXME

\subsubsection{Sample management and data containers.}
To represent the physical materials involved in a scientific process, we reuse the \textit{Set} ODP. The \texttt{aml:SampleSet} contains individual \texttt{aml:Sample} objects.
Crucially, AnIML allows for the definition of complex hierarchical relationships between samples.
We model this via the \texttt{aml:SampleInContainer} class, a situational pattern that links a \texttt{aml:Sample} to a physical \texttt{aml:Container} (a specific subclass of Sample), allowing the ontology to represent physical arrangements such as multi-well plates, racks, or trays.

To handle the heterogeneous data associated with samples (and other entities), we model AniML categories in the \texttt{aml:Category} class.
A Category acts as a (logical) flexible data container that can hold both Parameters and Series:
\begin{itemize}[topsep=1pt]
    \item \textbf{Parameters} are represented by the \texttt{aml:Parameter} class and describe single-value metadata fields (e.g., pH, temperature settings) as key-value pairs providing the context for the experimental entity they describe.
    \item \textbf{Series} are represented by \texttt{aml:SeriesSet} and \texttt{aml:Series}, these describe ordered and potentially multimodal array data (e.g., spectra, chromatograms, time-series). A \texttt{aml:Series} encapsulates the raw data points and is characterised by its dependency (independent variables like 'Time' vs. dependent variables like 'Absorbance') and data type.
\end{itemize}
Similarly to AnIML, \texttt{aml:Category} is reused throughout the ontology (e.g., in Results and Methods) to ensure a consistent pattern for data annotation.

\vspace{-1em} % FIXME

\subsubsection{Experimentation.}
The execution of experimental work is modelled through the \texttt{aml:Experiment} class.
Adopting the \textit{Sequence} ODP, an experiment is defined as an ordered list of \texttt{aml:ExperimentStep} objects.
Each step represents a discrete action or event and aggregates three distinct aspects of the scientific process:
\begin{enumerate}[topsep=1pt]
    \item \textbf{Method:} The \texttt{aml:Method} class describes \textit{how} the step is performed, including the procedural parameters and the \texttt{aml:Agent} (Software, Hardware, or Human) responsible for the action and its \texttt{aml:Role}.
    \item \textbf{Infrastructure:} The \texttt{aml:Infrastructure} class describes the environmental context and the physical setup used during the step. This allows for the separation of the \textit{procedure} (the method itself) from the specific \textit{assets} used to perform it (Infrastructure).
    \item \textbf{Result:} The \texttt{aml:Result} class encapsulates the output of the step, utilising the \texttt{aml:Category} pattern described above to store any data points or series.
\end{enumerate}

\vspace{-1em} % FIXME

\subsubsection{Provenance and audit trail.}
Regulatory compliance and versioning are handled by the \texttt{aml:AuditTrail}: a sequence of \texttt{aml:AuditTrailEntry} objects, each representing a versioned state of the document.
To track granularity, we introduce the \texttt{aml:Change} class, which links an entry to specific modifications (Creation, Deletion, or Modification).
Changes target objects that are subclasses of \texttt{aml:SignableItem} -- a superclass encompassing most critical entities (e.g., ExperimentSteps, Samples).
This mechanism extends to experimental actors -- linking \texttt{aml:Agent} to \texttt{aml:AuditTrail} to ensure that changes to the experimental environment (e.g. software, devices) are explicitly tracked.
Finally, the validity of these items is secured via \texttt{aml:Signature} objects within a \texttt{aml:SignatureSet}.

\begin{figure}[th]
    \centering
    \includegraphics[width=\linewidth]{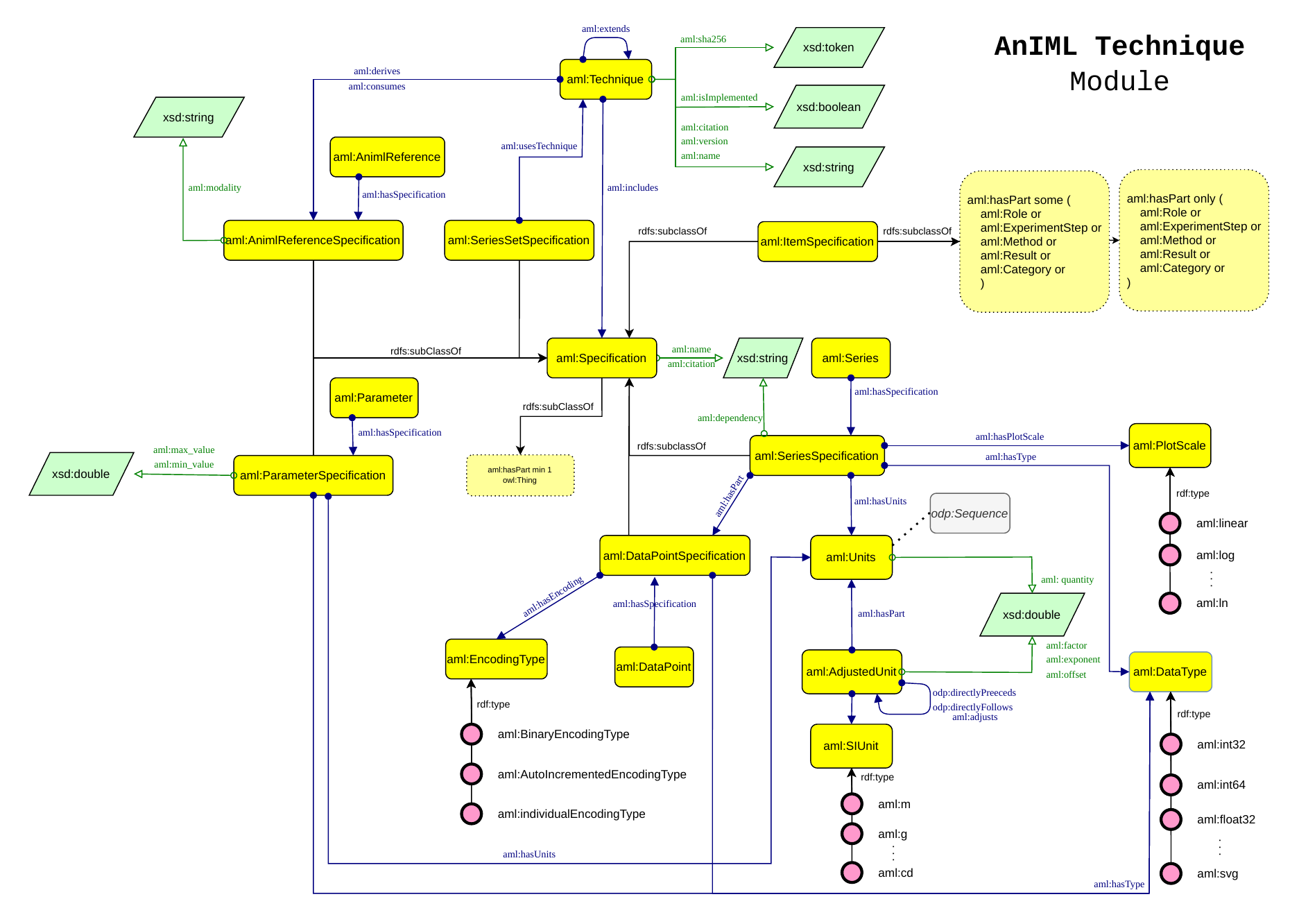}
    \caption{The AnIML technique module.}
    \label{fig:animl-technique}
\end{figure}

\subsection{Technique definitions}\label{ssec:animl-technique}

While the Core module provides the generic structure for data, it does not enforce domain-specific constraints (e.g., that a UV-Vis spectrum must have specific wavelength units).
In AnIML, this is handled by AnIML Technique Definition Documents (ATDDs).
We map this concept to the \texttt{aml:Technique} class, and illustrate this module in Figure~\ref{fig:animl-technique}.

\vspace{-1em} % FIXME

\subsubsection{Technique metadata.}
An \texttt{aml:Technique} entity captures the metadata describing the analytical method.
As shown in the top section of Figure~\ref{fig:animl-technique}, this includes the technique's \texttt{aml:name}, \texttt{aml:version}, and \texttt{aml:citation}.
We also model the implementation status via the boolean property \texttt{aml:isImplemented}.
Crucially, techniques in AnIML are modular and hierarchical; one technique file can build upon another.
We formalise this via the \texttt{aml:extends} object property, which allows a Technique to inherit specifications from a parent Technique.

\vspace{-1em} % FIXME

\subsubsection{Specifications and constraints.}
A Technique acts as a semantic blueprint.
An \texttt{aml:ExperimentStep} is linked to a Technique, which imposes constraints on the structure of that step.
Technique comprises various \texttt{aml:Specification} subclasses, defining the expected shape and data types of the experimental data:

\begin{itemize}
    \item \texttt{aml:ParameterSpecification} is used to constrain the allowed data types for parameters within a method or result. It dictates whether a specific parameter (e.g., ``Start Time'') must be a String, Float, Integer, or DateTime.
    \item \texttt{aml:SeriesSpecification} defines the requirements for data series (e.g., chromatograms). It constrains the \texttt{aml:Unit} (e.g., Hz, nm) and, importantly, the visualisation properties via \texttt{aml:PlotScale}. This ensures that a series is correctly interpreted as having a linear, logarithmic, or inverse scale.
    \item \texttt{aml:DataPointSpecification} provides low-level constraints on individual data points. As illustrated in Figure~\ref{fig:animl-technique}, it links to \texttt{aml:EncodingType}. This allows the ontology to specify how raw binary data streams (such as detector signals) should be parsed (e.g., \texttt{aml:BinaryEncodingType}).
    % ,\texttt{aml:AutoIncrementedEncodingType}).
    % \item \textbf{Reference Specification:} \texttt{aml:AnimlReferenceSpecification} constrains how steps link to other data, governing the valid targets for references within a specific technique.
\end{itemize}

\subsection{The AnIML reference pattern}\label{ssec:animl-reference-odp}

\begin{figure}[th]
    \centering
    \includegraphics[width=0.7\linewidth]{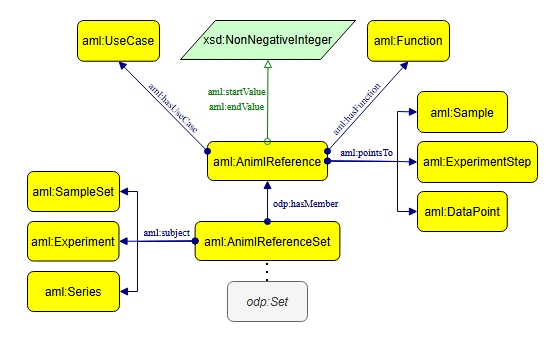}
    \caption{The AnIML reference pattern.}
    \label{fig:reference_pattern}
\end{figure}

A significant challenge in the XML schema is the use of implicit ID/IDREF mechanisms to link data (e.g., a Result referencing a Sample ID).
These links are opaque to standard semantic reasoners.
To address this, we introduce the \textit{AnIML Reference Pattern}, depicted in Figure~\ref{fig:reference_pattern}.

We promote these pointers into the \texttt{aml:AnimlReference} class.
This class explicitly models the relationship using the \texttt{aml:pointsTo} object property, which can target a \texttt{aml:Sample}, \texttt{aml:ExperimentStep}, or \texttt{aml:DataPoint}.
Beyond simple linking, the pattern captures the context of the reference:
\begin{itemize}
    \item \textbf{Role:} the \texttt{aml:hasFunction} property describes why the reference exists (e.g., ``reference blank'' vs ``analyte'') or its role within the reference set.
    \item \textbf{Scope:} for array data, \texttt{aml:startValue} and \texttt{aml:endValue} properties allow referencing a specific subset or slice of a data series.
\end{itemize}

References are grouped into a \texttt{aml:AnimlReferenceSet}.
To ensure valid scoping, this set is anchored to a specific container via the \texttt{aml:subject} property, which restricts the reference context to a specific \texttt{aml:SampleSet}, \texttt{aml:Series} or \texttt{aml:Experiment}.
We employ SHACL shapes to validate that the subject of the reference set aligns with the targets of the contained references.
% (Listing~\ref{lst:shacl_animl}).

Although originating as a Reengineering Pattern to resolve XML's ID/IDREF mechanism, we classify this construct as a Content ODP rather than an ontology idiom~\cite{falbo2013ontology}.
The distinguishing criterion is whether the pattern would persist under a change of encoding language~\cite{falbo2013ontology}: an idiom would simply create a flat \texttt{owl:ObjectProperty} to mimic the XML pointer, whereas the AnIML Reference Pattern promotes references to first-class
entities that carry their own function (\texttt{aml:hasFunction}), use
case (\texttt{aml:hasUseCase}), and array scope (\texttt{aml:startValue},
\texttt{aml:endValue}), bound to typed containers via
\texttt{aml:subject}.
These concepts capture domain knowledge about how references operate in analytical experimentation and would be required regardless of the encoding language.
% To illustrate the application of this ODP for data slicing, Listing~\ref{lst:cq96} presents the SPARQL query for retrieving the start index of a value set (CQ-96).
%% \input{figures/pattern_shacl}
% \input{figures/pattern_cq96.tex}
%
To illustrate, Listing~\ref{lst:shacl-ref} shows the SHACL shape that enforces scoping constraints on \texttt{aml:AnimlReferenceSet}. The shape ensures that when the subject of a reference set is a \texttt{SampleSet}, the contained references may only point to \texttt{Sample} instances, preventing, for example, a sample-scoped reference set from accidentally targeting experiment steps.

\begin{lstlisting}[caption={SHACL shape enforcing \texttt{aml:AnimlReferenceSet} scoping constraints.},label={lst:shacl-ref},language=SPARQL,basicstyle=\ttfamily\small,breaklines=true]
aml:AnimlReferenceSetShape a sh:NodeShape ;
  sh:targetClass aml:AnimlReferenceSet ;
  sh:or ( aml:SampleSetConfiguration
          aml:ExperimentConfiguration
          aml:SeriesConfiguration ) ;
  sh:message "Subject type not match reference target." .

aml:SampleSetConfiguration a sh:NodeShape ;
  sh:property [ sh:path aml:subject ;
                sh:class aml:SampleSet ; sh:minCount 1 ] ;
  sh:property [ sh:path odp:hasMember ;
    sh:node [ a sh:NodeShape ;
      sh:property [ sh:path aml:pointsTo ;
                    sh:class aml:Sample ] ] ] .
\end{lstlisting}
\subsection{Ontology alignment}\label{ssec:oa-results}

To facilitate cross-standard interoperability, we generated ontology alignments with the Allotrope Foundation Ontologies (AFO), a widely adopted standard for laboratory data management~\cite{stromert2022ontologies4chem}.
The AFO also provide a standard vocabulary for describing equipment, materials, processes, and results, and are in turn aligned, at the top-level, with the Basic Formal Ontology (BFO) \cite{otte2022bfo}.
% \elliott{By aligning the AnIML Ontology to Allotrope, we ensure that the ontology has a sufficient scope which enables interoperability, but still retains the stand-alone purpose of the AnIML schema}

Because AFO and AnIML differ in scope and modelling granularity, alignment coverage was deliberately scoped rather than exhaustive.
While AFO provides a broad, BFO-aligned vocabulary for laboratory processes, AnIML strictly specialises in analytical instrument data.
Consequently, full concept-level equivalence is neither achievable nor an appropriate evaluation metric.
Instead, we prioritise purposive coverage: ensuring that AnIML data types faithfully map to AFO counterparts to support cross-standard interoperability.

Alignment candidates are first generated using state-of-the-art ontology alignment techniques including LogMap \cite{jimenez2011logmap}, ontology embeddings \cite{he2022bertmap}, neural-based methods from DeepOnto \cite{he2024deeponto} and OntoAligner \cite{babaei2025ontoaligner}.
All alignments are released using the Simple Standard for Sharing Ontological Mappings (SSSOM) \cite{matentzoglu2022simple}, a format that provides relevant metadata to describe each mapping.

% \begin{wraptable}{h}{0.4\columnwidth}
%     \centering
%     \footnotesize
%     \caption{Alignment stats.}
%     \label{tab:alignments}
%     \begin{tabular}{@{}lrrr@{}}
%         \toprule
%         \textbf{Method} & \textbf{\#} & \textbf{\checkmark} & \textbf{(\%)} \\ \midrule
%         LogMap   & 76  & 44 & 58  \\
%         BERTmap  & 35  & 19 & 54  \\
%         DistMult & 18  & 12 & 67  \\
%         CompGCN  & 14  & 11 & 79  \\
%         TransD   & 13  & 11 & 85  \\
%         ConvE    & 11  & 10 & 91  \\
%         TransE   & 10  & 10 & 100 \\ \midrule
%         \textbf{Total} & \textbf{177} & \textbf{117} & \textbf{66} \\ \bottomrule
%     \end{tabular}
% \end{wraptable}

% All mappings were manually validated
% \elliott{and scoped coverage was evaluated to ensure ... . Coverage of the mappings was favoured against completeness due to the differences of modelling granularity, scope, and overlap between concepts in both AnIML and Allotrope respectively. 
% For example, the ... in AnIML represents ..., but this is not found in Allotrope and does not appear in the alignments.}
Since all methods natively produce equivalence mappings (\texttt{owl:equivalentClass}), we manually reviewed and relaxed these relationships using SKOS~\cite{miles2009skos} whenever a looser semantic correspondence was identified.
In total, the ensemble of methods generated $176$ candidate alignments (including duplicates).
Following curation, $121$ were accepted as strict equivalences (\texttt{owl:equivalentClass}); and $20$ mappings were manually redefined into suitable SKOS variants: \texttt{skos:relatedMatch} ($11$), \texttt{skos:narrowMatch} ($5$), and \texttt{skos:broadMatch} ($4$)
% \elliott{An example of this is the mapping between \texttt{animl/result} and Allotrope \texttt{result\#AFR\_0000207} (Assay result). BERTmap initially mapped this as an equivalent mapping, before skos redefinition through into \texttt{skos:broadMatch} through manual validation}.
The acceptance rates for equivalent mappings varied by method: LogMap ($64\%$, $49$), BERTmap ($54\%$, $19$), ConvE ($91\%$, $10$), TransE ($100\%$, $10$), TransD ($85\%$, $11$), CompGCN ($79\%$, $11$), and DistMult ($67\%$, $12$).
Overall, this process yielded $47$ unique alignments.

Beyond SKOS, certain structural mismatches require specific redefinitions.
For instance, the alignment between AnIML \texttt{Method} and Allotrope \texttt{Method name} is redefined using a \texttt{partOf} relationship.
Similarly, LogMap initially mapped the property \texttt{animl/hasUnits} to the class \texttt{qudt\#unit}; this was manually corrected to link the class \texttt{animl/Units} to \texttt{qudt\#unit} via \texttt{owl:equivalentClass}.

\section{Evaluation}\label{sec:validation}
We evaluate the AnIML Ontology through two complementary strategies:
\emph{verification}, confirming that the ontology is structurally sound
and can answer the questions it was designed for; and \emph{validation},
checking that it excludes unintended states of affairs. We address each
in turn.

\noindent\textbf{Structural and logical verification.}
The core ontology and its AFO-aligned version were evaluated using the
OntOlogy Pitfall Scanner! (OOPS!)~\cite{poveda2014oops} to detect
structural anomalies and potential logical inconsistencies. The resulting
warnings were systematically logged and addressed through minor
refinements to the ontology's axioms, metadata, and cross-references,
ensuring a robust foundational structure prior to data instantiation.

\noindent\textbf{KG population and alignment testing.}
A dataset of 10 real-world AnIML files provided by Unilever, obfuscated
to protect proprietary information, was used to populate the ontology. This transformation acted as a feedback loop during ontology
engineering: where the ontology failed to capture underlying data
structures, the model was refined in subsequent iterations. The
translation of the final test set was successful, demonstrating that the
ontology can express complex industrial experimental records. To evaluate
alignment soundness (c.f.\ Section~\ref{ssec:oa-results}), we expressed
the AnIML examples through our AFO mappings; a semantic reasoner
confirmed logical consistency across all the instances.

\noindent\textbf{CQ verification.}
We tested the ontology against a representative subset of 40 requirements
elicited in Section~\ref{ssec:requirement-engineering} through CQ
verification~\cite{blomqvist2012ontology}. This involves formalising the
selected CQs into SPARQL queries; the successful formulation of these
queries confirms that the ontology contains the necessary semantic
constructs to answer the core domain questions defined by the experts.
The full set of SPARQL queries is documented in the repository.
To illustrate, Listing~\ref{lst:sparql-cq96} shows the SPARQL query for CQ-96 (``What is the start index for a specific value set in a series?''), which retrieves data slice boundaries using the AnIML Reference Pattern. This query would not be expressible against the original XML schema, where the ID/IDREF mechanism lacks the explicit scoping semantics captured by \texttt{aml:startValue}.

\begin{lstlisting}[caption={SPARQL query for CQ-96, retrieving data slice boundaries via the AnIML Reference Pattern.},label={lst:sparql-cq96},language=SPARQL,basicstyle=\ttfamily\small,breaklines=true]
PREFIX aml: <http://www.w3id.org/animl/ontology/>
SELECT ?reference ?series ?startIndex
WHERE {
  ?reference a aml:AnimlReference ;
    aml:pointsTo ?series ;
    aml:startValue ?startIndex .
  ?series a aml:Series .
}
\end{lstlisting}

\noindent\textbf{Validation via adversarial competency questions.}
\label{sec:validation_adversarial}
CQ verification confirms that the ontology \emph{can} answer the
questions it was designed for, i.e. that intended states of affairs are
representable. However, verification alone cannot guarantee that the
ontology excludes \emph{unintended} states of affairs: models that are
logically valid under OWL semantics yet represent situations that should
not occur in the domain~\cite{sales2015antipatterns}. To address this
gap, we perform a validation step grounded in the notion of ontological
anti-patterns.

Sales and Guizzardi~\cite{sales2015antipatterns} identify six recurrent
anti-patterns: error-prone modelling structures that typically cause
the set of valid model instances to diverge from the set of intended
ones. Their catalogue was developed for OntoUML, whose meta-model is
grounded in UFO. Since the AnIML Ontology reuses DOLCE-based ODPs and is
encoded in OWL~2, we cannot directly apply the OntoUML approach. Instead,
we adapt the conceptual framework by systematically generating
\emph{adversarial negative competency questions}, i.e. requirements that the
ontology must \emph{not} satisfy, and enforcing them via SHACL
constraints that can be mechanically checked against instance data.

\noindent\textit{Protocol.}
Starting from the 102 positive CQs, we generate candidate negative CQs
targeting specific anti-pattern families: type confusion and disjointness
failures (BinOver), role--type incompatibility in mediated structures
(RelOver), hidden subtype-level constraints (ImpAbs), ordering and
dependency violations between related associations (RelSpec), identity
uniqueness (RepRel), and structural cycles (AssCyc). Each candidate was
reviewed by two ontology engineers to confirm that (i)~it captures a
genuinely unintended state of affairs in the AnIML domain, and (ii)~it
maps to a recognisable anti-pattern category. This process yielded 20
validated negative requirements, summarised in Table~\ref{tab:negcqs}.

\input{figures/adversarial_CQs_tablev3}

\noindent\textit{Enforcement and testing.}
Each negative requirement is formalised as a SHACL shape, predominantly
using SPARQL-based constraints to express conditions that go beyond what
OWL axioms alone can enforce under the open-world assumption. For every
shape, we provide two companion test datasets: a \emph{positive} dataset
containing only intended instances (expected to pass validation) and a
\emph{negative} dataset introducing a targeted violation (expected to
trigger the constraint). For example, negative requirement~17
(Table~\ref{tab:negcqs}) enforces that an experiment step's
infrastructure may not reference steps occurring later in the
experimental sequence --- an unintended state of affairs where the
infrastructure-reference association violates the temporal ordering of
steps (RelSpec). The SHACL shape uses a transitive property path over
\texttt{seq:directlyPrecedes} to detect forward references across
arbitrarily long step chains. Similarly, requirement~6 prevents a
\texttt{HardwareAgent} from filling an \texttt{Author} role,
constraining an otherwise permissive mediation between overlapping agent
subtypes (RelOver).

\noindent\textit{Scope and limitations.}
This validation targets the most likely sources of unintended models given
the AnIML domain structure, rather than claiming exhaustive coverage of
all possible anti-pattern instances. The 20 negative requirements
concentrate on the core and technique modules where the risk of
misinterpretation is highest, particularly around agent--role assignments,
experiment step ordering, and reference scoping. Future work will extend
coverage as the ontology evolves and additional stakeholder data becomes
available.

%% file: figures/adversarial_CQs_tablev3.tex
\begin{table}[t!]
\begin{threeparttable}
\caption{Adversarial competency questions mapped to ontological
anti-pattern categories~\cite{sales2015antipatterns}. AP abbreviations:
AssCyc = Association Cycle; BinOver = Binary Relation Between Overlapping
Types; RelOver = Relator Mediating Overlapping Types; RelSpec = Relation
Specialization; ImpAbs = Imprecise Abstraction; RepRel = Repeatable
Relator Instances. All SHACL shapes and test data are in the repository.}
\label{tab:negcqs}
\centering
\setlength{\tabcolsep}{3pt}
\renewcommand{\arraystretch}{1.25}
\scriptsize
\rowcolors{2}{gray!8}{}%
\begin{tabularx}{\textwidth}{@{}r X l l@{}}
\toprule
\rowcolor{white}%
\# & Negative Requirement & AP & Mechanism \\
\midrule
1  & No cycles in \texttt{directlyPrecedes}/\texttt{directlyFollows}
   & AssCyc & Cycle detection \\
2  & Nothing may both directly follow and directly precede anything else
   & AssCyc & Mutual-inverse check \\
3  & A reference may not point to a member unless its reference set points to the containing structure
   & RelSpec & Scope alignment \\
4  & A \texttt{HumanAgent} may not have an OS or version
   & BinOver & Property exclusion \\
5  & A \texttt{HardwareAgent} may not have a phone or email
   & BinOver & Property exclusion \\
6  & A \texttt{HardwareAgent} may not be assigned an author role\tnote{a}
   & RelOver & Role--type restriction \\
7  & A \texttt{SeriesSet} spec.\ may not include non-\texttt{SeriesSpecification} items
   & ImpAbs & Type restriction \\
8  & A \texttt{SeriesSpec.}\ may not have ${>}$1 \texttt{DataPointSpec.}\ unless exactly 2 (AutoIncr.\ + Individual)
   & ImpAbs & Cardinality + type \\
9  & A required \texttt{AnimlRefSpec.}\ may not exist without a concrete reference
   & RelSpec & Conditional existence \\
10 & No role may be required without also being the role of an agent\tnote{b}
   & ImpAbs & Completeness check \\
11 & A container without \texttt{SampleInContainer} may not have any type other than Simple
   & ImpAbs & Conditional type \\
12 & An Increment \texttt{DataPoint} may not have a non-Float value
   & ImpAbs & Datatype restriction \\
13 & A \texttt{BinaryEncoding} \texttt{DataPoint} may not have a non-Binary value
   & ImpAbs & Datatype restriction \\
14 & Two or more \texttt{aml:id}s may not share the same value
   & RepRel & Uniqueness check \\
15 & Parameter min/max may not be non-null unless type is \texttt{Int} or \texttt{Float}
   & ImpAbs & Conditional datatype \\
16 & No parameter value may fall outside its spec's min/max range
   & ImpAbs & Range check \\
17 & Infrastructure may not reference \texttt{ExperimentStep}s after the linked step\tnote{c}
   & RelSpec & Ordering check \\
18 & Infrastructure may not reference samples consumed by earlier steps
   & RelSpec & Temporal dependency \\
19 & Results may not reference samples already derived by another step
   & RelSpec & Temporal dependency \\
20 & Infrastructure/results may not cross-reference samples derived/consumed by its step
   & RelOver & Cross-ref.\ exclusion \\
\bottomrule
\end{tabularx}
\begin{tablenotes}[flushleft]\scriptsize
\item[a] \texttt{Role} mediates between \texttt{Agent} subtypes; without this constraint the overlapping type hierarchy permits incompatible role assignments.
\item[b] \texttt{requiresRole} hides the constraint that role assignment implies agent allocation.
\item[c] Infrastructure-reference and temporal ordering associations must be mutually constrained.
\end{tablenotes}
\end{threeparttable}
\end{table}

%% file: sections/discussion_conclusions.tex
\section{Discussion and conclusions}

This paper introduces the AnIML Ontology, an OWL\,2 ontology that
formalises the Analytical Information Markup Language, bridging XML-based
experimental records and the Semantic Web.
The ontology was developed via an expert-in-the-loop approach combining LLM-assisted requirement elicitation with collaborative ontology engineering, and evaluated through structural verification, data-driven KG population, CQ verification, and a novel validation protocol using adversarial negative competency questions mapped to ontological anti-patterns and enforced via SHACL.

The current version (1.1.0) maintains close compatibility with the AnIML
XML schema, simplifying the transition of legacy data into knowledge
graphs. This creates a learning curve for users already familiar with
AnIML, who can start by reusing the core ontology and extend it based on
their own data, environments, and use cases, for example, by adding
richer metadata, introducing new relationships across samples, or
registering new technique definitions.
Future versions will prioritise reuse of established ontologies for processes~\cite{carriero2025procedural}, agents~\cite{malone2014software}, units~\cite{hodgson2011qudt}, and provenance~\cite{moreau2011open}.
Beyond Allotrope, further
alignments with ontologies from the OBO Foundry~\cite{smith2007obo} and
BioPortal~\cite{rubin2008bioportal}, e.g. 
ChEBI~\cite{dematos2010chebi} and MDO~\cite{li2020ontology}, would
bridge experimental metadata with domain ontologies for chemical
characterisation and material discovery.
The adoption of AnIML by companies such as Unilever, GSK, and Merck KGaA, and consortia like SiLA~\cite{bar2012sila}, underscores the need for such integration.